\documentclass[10pt,twocolumn,letterpaper]{article}

\usepackage{iccv}
\usepackage{times}
\usepackage{epsfig}
\usepackage{graphicx}
\usepackage{amsmath}
\usepackage{amssymb}
\usepackage{cite}
\usepackage{dsfont}
\usepackage{subfigure}

\usepackage{sidecap}
\usepackage{wrapfig}

\usepackage[pagebackref=true,breaklinks=true,letterpaper=true,colorlinks,bookmarks=false]{hyperref}

\iccvfinalcopy 

\def\iccvPaperID{2758} 

\ificcvfinal\fi
\begin{document}

\title{Customizing First Person Image Through Desired Actions}

\author{Shan Su\\
University of Pennsylvania\\
{\tt\small sushan@seas.upenn.edu}
\and
Jianbo Shi\\
University of Pennsylvania\\
{\tt\small jshi@seas.upenn.edu }
\and
Hyun Soo Park\\
University of Minnesota\\
{\tt\small hspark@umn.edu}
}

\maketitle

\begin{abstract}

This paper studies a problem of inverse visual path planning: creating a visual scene from a first person action. Our conjecture is that the spatial arrangement of a first person visual scene is deployed to afford an action, and therefore, the action can be inversely used to synthesize a new scene such that the action is feasible. As a proof-of-concept, we focus on linking visual experiences induced by walking. 

A key innovation of this paper is a concept of ActionTunnel---a 3D virtual tunnel along the future trajectory encoding what the wearer will visually experience as moving into the scene. This connects two distinctive first person images through similar walking paths. Our method takes a first person image with a user defined future trajectory and outputs a new image that can afford the future motion. The image is created by combining present and future ActionTunnels in 3D where the missing pixels in adjoining area are computed by a generative adversarial network. Our work can provide a travel across different first person experiences in diverse real world scenes.




\end{abstract}

\section{Introduction}

\begin{figure}[t]
  \centering  
  \label{Fig:teaser}\includegraphics[width=0.47\textwidth]{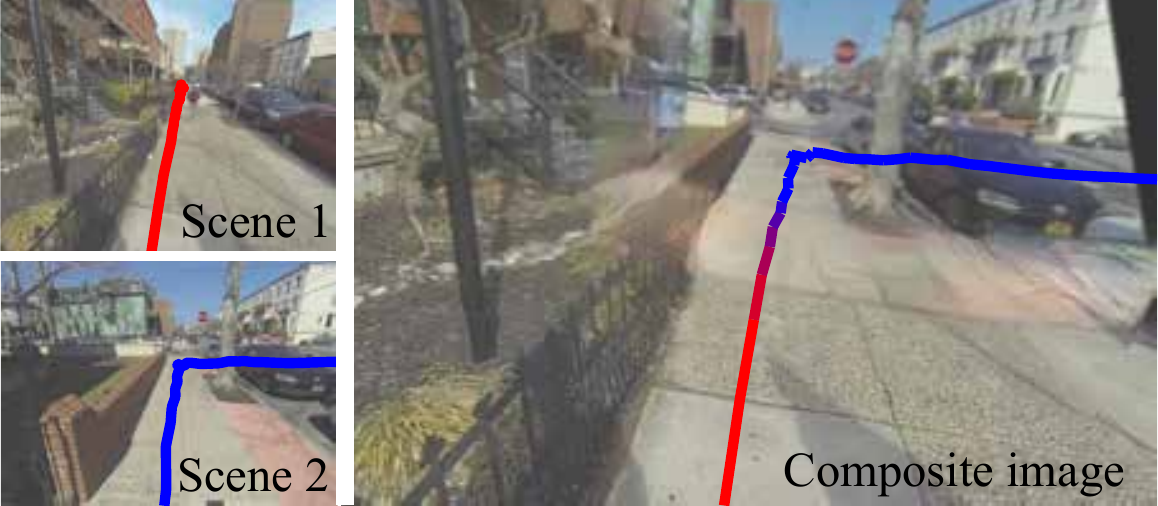}
  \caption{This paper presents a method to connect two first person images through their future action. Scene 1 and 2 are synthesized to produce a novel image (right) that affords a right turn action at the intersection. The trajectories represent the future walking path on the ground where the color encodes the transition.} 
\end{figure}

Imagine you have a picture of the scene in front of you, as shown in Scene 1 of Figure~\ref{Fig:teaser}. You see the road, the parked cars (right), and fence (left).  This scene makes you move straight.  What if you want to make a right turn half way down the road. How do you {\em imagine} the scene so that you can modify the action\footnote{Fran\c cois Vogel's art work illustrates visual experiences created by actions.}?  We would need to create an intersection for you to turn, and a side street to continue onto. In another example, if we want to enter a building to the left, we would need to create an entrance on the building.
 
In this paper, we propose a problem of inverse visual path planning, i.e., creating a visual scene from a future path.
Our key insight is based on Gibson's ecological perception~\cite{gibson:1979}: a strong duality between  visual scene and action. The spatial arrangement of first person scenes affords actions, and therefore, it is possible to visualize what we would see based on {\em how we will act}. A key challenge is to embed visual semantics of a new future action into the present scene, i.e., synthesizing pixels such that the present image can afford future as shown in Composite image of Figure~\ref{Fig:teaser}. 

We address this challenge by leveraging a novel concept of {\em ActionTunnel}---a 3D virtual tunnel along the future trajectory encoding what the camera wearer will visually experience as moving into the scene akin to the spacetime tunnel in physics. ActionTunnel connects two distinctive scenes in a spatially consistent way, which allows creating a new image that affords the desired future action. As a proof-of-concept, we focus on linking such visual sensations through walking. 

Our method takes a first person RGBD image with a user defined future trajectory and outputs a new image that the desired motion is feasible. We represent an image using ActionTunnel along the future trajectory where its cross section is determined by the walkable pixels on the ground plane. The future visual scene is selected according to the similarity of future trajectories and their surrounding visual context. We combine present and future ActionTunnels in 3D to generate a novel image where the missing pixels in adjoining area are computed by a generative adversarial network~\cite{goodfellow:2014}. The resulting ActionTunnel is projected onto the camera pose, which produces a continuous transition.



We cast this problem as learning visual context of walking path memory. A visual scene can be decomposed according to short and long term desired paths where we find their transition and geometric alignment of the present and future scenes, and glue them back into a composite image.

\noindent\textbf{Why useful?} An ability to connect visual sensations is a key design factor of AI because it can index visual scenes in a spatially persistent way. For instance, our visual sensation alignment can generate the contents for virtual reality (VR), providing new supervisionary signals for learning visual semantics across time, and share virtual experiences via first person videos. Also this can close the loop of the planning by perception paradigm, i.e., our work provides perception through planning, which can feedback to the planning. Beyond AI and VR, this work brings a new opportunity to study the human visual memory system, establishing a tight integration of memory into visual perception and motor skill~\cite{wolpert:2011}.

\noindent\textbf{Contribution} To our best knowledge, this is the first paper that addresses connecting first person visual sensations through actions. The core technical contributions include: 1) ActionTunnel representation: we develop a novel visual representation persistent to the 3D spatial structure given future trajectory. 2) Visual sensation alignment and retrieval: our algorithm predicts a scene that produces a plausible transition by aligning ActionTunnel. 3) Gap filling: we use a generative adversarial network to fill the pixels in adjoining area, which connects the ActionTunnel. We demonstrate that our work can provide travel across different first person experiences in diverse real world scenes.

\section{Related Work}

\begin{figure*}[th]
  \centering  
      \subfigure[Geometry]{\label{Fig:geometry}\includegraphics[height=0.17\textheight]{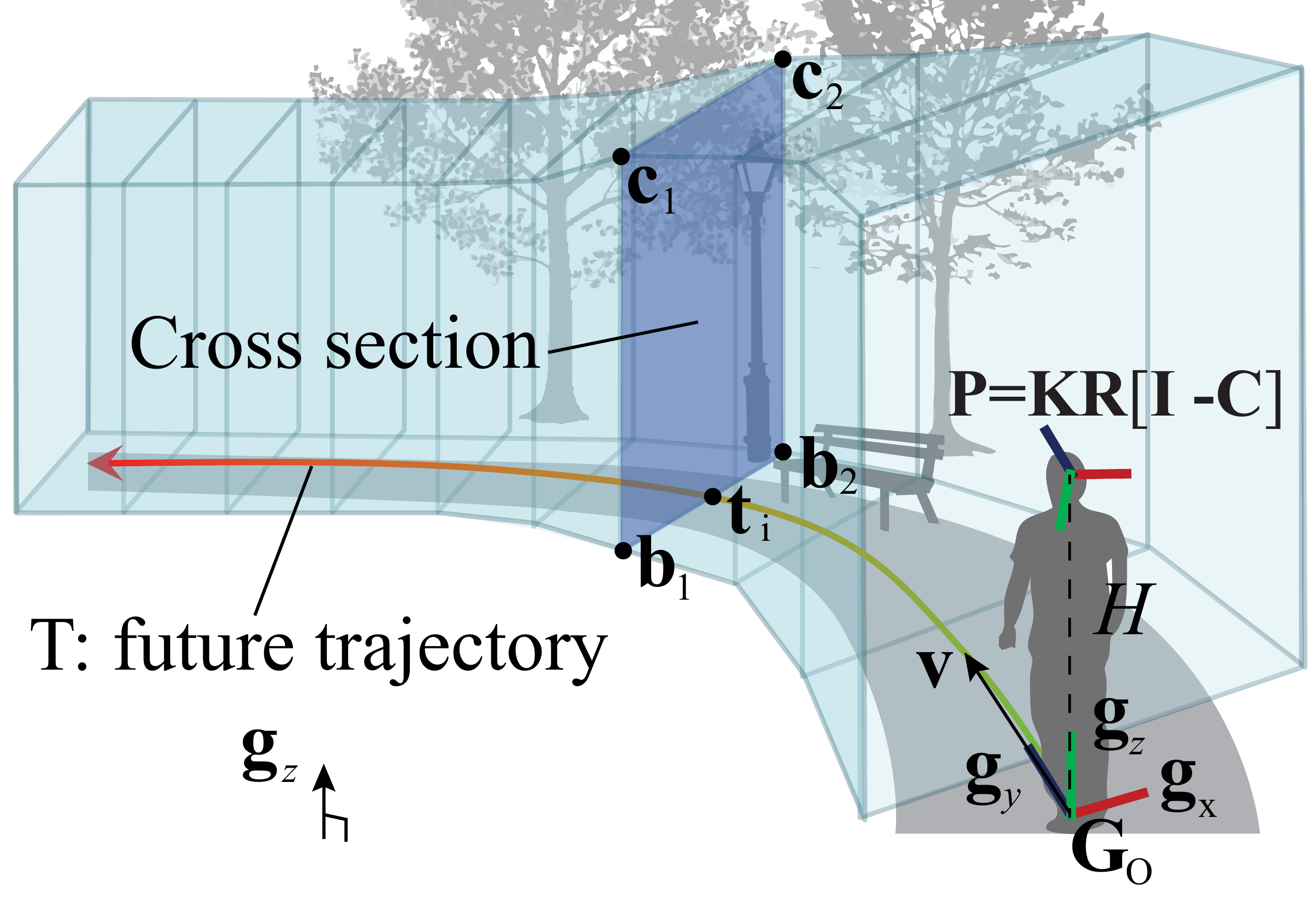}}~
      \subfigure[Rectified image]{\label{Fig:rectify}\includegraphics[height=0.17\textheight]{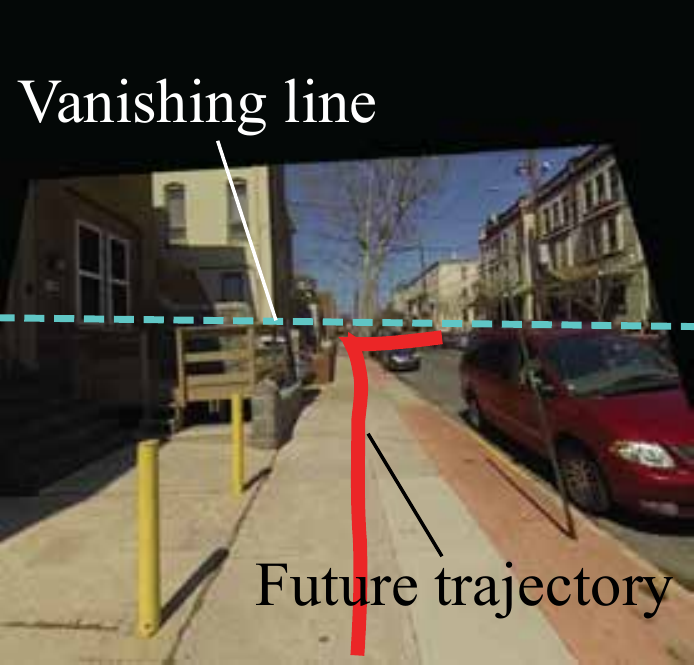}}~
      \subfigure[ActionTunnel]{\label{Fig:actiontunnel}\includegraphics[height=0.17\textheight]{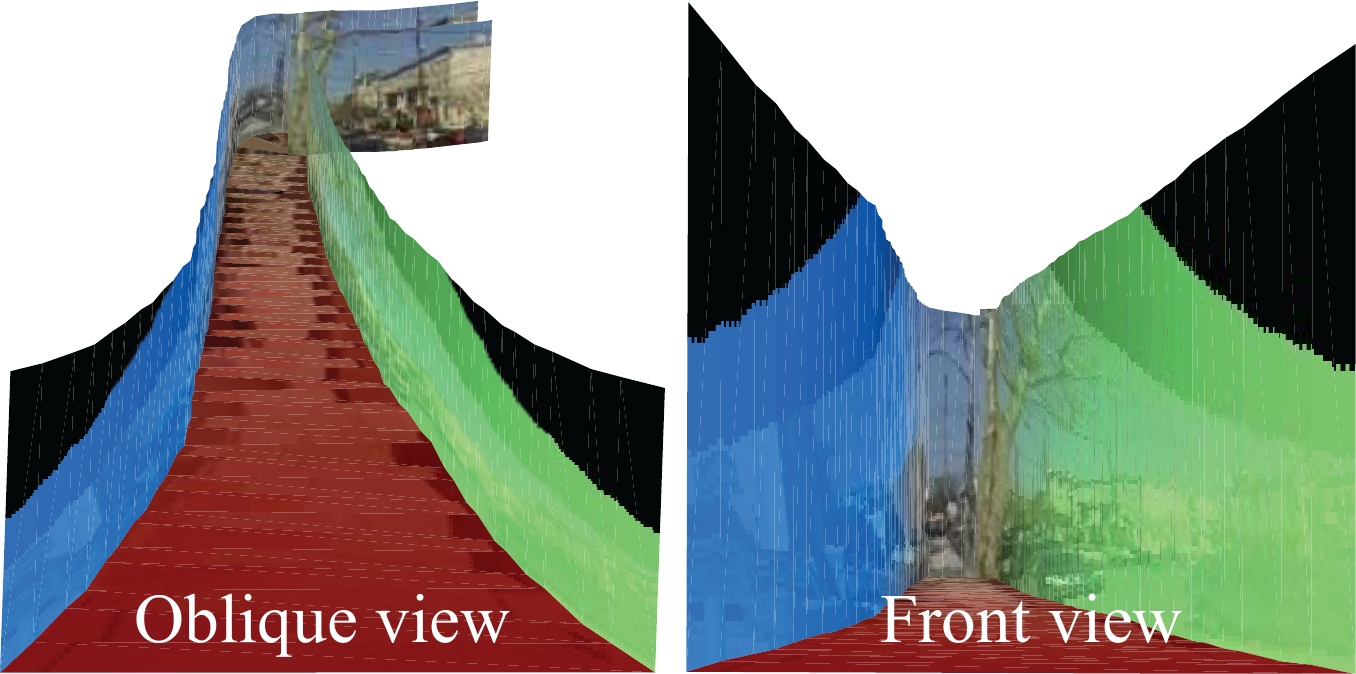}}~~
  \caption{(a) ActionTunnel is a 3D virtual tunnel along the future trajectory that encodes what the camera wearer will visually experience as moving into the scene. (b) We rectify all first person images such that the vanishing lines are aligned, which allows learning the visual representation of a first person image efficiently by eliminating severe 3D head motion. (c) ActionTunnel represented in the 3D proxemic space (log-cylindrical coordinate) is constructed along the future trajectory. The color and transparency represent the surface type and time, respectively.} 
  \label{Fig:action11}
\end{figure*}

Connecting human visual sensations through memory system, i.e., how we index and store what we have seen, has been the central theme of neuroscience~\cite{ungerleider:1998,vogel:2004} and cognitive vision science~\cite{melcher:2001,reinhart:2015,brady:2011}. This has motivated computer vision and graphics research to build a computational visual memory system that connects the humongous image and video data in a semantically meaningful way.

A core machinery behind linking the visual sensations is the space and time registration of visual data, discovering invariant visual patterns that can relate distinctive instances/experiences~\cite{gibson:1979}. Due to its highly structured shape, human faces are one of the earliest targets in visual registrations in computer vision (e.g., eigenface~\cite{turk:1991}). Such face alignment has been the backbone of reconstructing occluded and noisy face~\cite{wright:2011}, tracking~\cite{saragih:2011} and re-targeting facial expressions~\cite{shlizerman:2010}, and building a progression of a photo album via spatiotemporal registration~\cite{shlizerman:2011}. Beyond a single object such as a face, aligning spatial context together can strengthen the links between visual sensations~\cite{oliva:2007,malisiewicz:2009}. Such contextual visual information is deeply embedded in our daily lives, which provides a strong cue to localize the image in the world~\cite{shrivastava:2011,doersch:2012}, recognize human sketches~\cite{lee:2011}, explore multiple images together~\cite{zhu:2014}, and arrange images based on space and time~\cite{lee:2013}. Notably, 3D spatial registration using structure from motion links a set of unorganized images from different perspectives~\cite{snavely:2006, snavely:2008} or videos~\cite{tompkin:2012, arev:2014}. Also precise temporal ordering based on visual semantics produces consistent visual experiences. Moving objects in multiple images provides a geometric constraint across time~\cite{PhotoSeq_ECCV12}, finding a smooth transition between images based on geometry and appearance generates a time-lapse video~\cite{brualla:2015}, and a prior knowledge of urban scene structure such as buildings relates historical pictures~\cite{schindler:2010}.

Unfortunately, such visual registrations are still not sufficient to describe the way we connect our visual episodic memory. It mainly focuses on {\em knowing} a factual event, e.g., where it was, how it looked, and what happened, while our visual memory system delves for {\em remembering} the personal experiences through interactions, e.g., what I did and how I felt~\cite{tulving:1983}. We attempt to fill this gap by exploiting first person videos that capture the visual sensation induced by walking actions.

A key property of a first person video is that it can closely capture what is important to the camera wearer through {\em interactions}~\cite{lee:2012,bertasius:2016}, which is highly relevant to episodic visual memories. Two visual signals delineate the first person interactions. First, the actions of the camera wearers link past experiences. The gaze direction tells us about the visual attention while interacting with objects~\cite{li:2013,fathi:2011} and people\cite{fathi:2012,park:2012,park:2013}, and hand manipulation provides a strong cue to recognize activities~\cite{fathi:2011,rogez:2016,ma:2016,li:2015}. Therefore, it is possible to infer what they did, and further identify how much they paid attention~\cite{su_grauman:2016}, which physical force they experienced~\cite{park:2016_physics}, and how they felt through physiological state estimation~\cite{hernandez:2014}. Second, the surrounding visual scenes relates visual sensations similar to D\'ej\`a vu experiences. First person visual data are highly structured due to camera placement, anthropometric constraints, and proxemics~\cite{hall:1963}, which allows characterizing visual spatial layout. For instance, unique dynamics of sport scenes can be recognized~\cite{kitani:2011}, the second person's activities with respect to the first person can be inferred~\cite{ryoo:2013}, and semantic segmentation for body parts can be used to understand the state of the interacting objects~\cite{li:2013, li_cvpr:2013}. Using such first person characteristics, the future actions can be forecasted via behavioral cloning~\cite{park:2016_future,singh:2016} and visual features can be learned without supervision~\cite{jayaraman:2015,agrawal:2015}. These two signals from first person videos are integrated to summarize lifelogging videos~\cite{zhang:2016, lu:2013}, identifying an memorable moment~\cite{xiong:2014}, editing first person videos~\cite{kopf:2014,arev:2014}.

The innovation of our work is to use first person actions as a cue to link between visual sensations without supervision by aligning them through ActionTunnel. This produces a continuous transition from one scene to the other. 

\section{ActionTunnel Model}\label{sec:actiontunnel}

We construct an ActionTunnel model---a 3D virtual tunnel along the future trajectory encoding what the camera wearer will visually experience as moving into the scene (Figure~\ref{Fig:geometry}). This model allows transitioning from one visual scene to the other according to the action (Section~\ref{Sec:connection}). The desired property of the ActionTunnel is the ability \textbf{to edit the present image to embed the visual semantics of the desired future action without manual supervision}. We represent ActionTunnel in the 3D proxemic space~\cite{hall:1963} encoding visual and spatial semantics that affords the walking actions. 




\subsection{3D Proxemic Space} \label{Sec:fps}
We define a 3D proxemic space using a log-cylindrical coordinate that highly emphasizes on the short range area. A 3D point $\mathbf{x}=\mathbf{G}_{o}+X\mathbf{g}_x+Y\mathbf{g}_y+Z\mathbf{g}_z$ in the world coordinate is mapped to $(r, \theta, h)$ such that $r=\log{\rho}$, $\theta=\text{atan2}(Y, X)$, and $h=Y/\rho$ where $\rho=\sqrt{X^2+Y^2}$, $\mathbf{g}_x, \mathbf{g}_y \in \mathds{S}^2$ are two unit vectors spanning the ground plane, and $\mathbf{g}_z = \mathbf{g}_x \times \mathbf{g}_y$ is its surface normal as shown in Figure~\ref{Fig:geometry}. We align $\mathbf{g}_y$ with the instantaneous velocity, $\mathbf{v}=\mathbf{C}_{t+1}-\mathbf{C}_t$, $\mathbf{C}\in \mathds{R}^3$ is the optical center of the first person camera, and $\mathbf{G}_o=\mathbf{C}-H\mathbf{g}_z$ is the origin of the ground plane where $H$ is the height of the camera wearer.

This proxemic space preserves 3D distance near the camera wearer. The EgoRetinal map~\cite{park:2016_future} (Figure~\ref{Fig:ego1})\footnote{The EgoRetinal map is a visual representation experienced from first person view but visualized in an overhead bird-eye map, akin to an illustrated tourist map.} is a special instantiation of this log-cylindrical space, which is its ground plane projection. This reflects both first person coordinate and 3D world coordinate: it does not introduce significant pixel perspective distortion while preserving 3D visual sensation similar to EgoRetinal map.

Note that the camera orientation is not necessarily aligned with the moving direction. In order to learn a first person visual representation invariant to head movement and camera placement, we rectify the camera orientation $\mathbf{R}\in SO(3)$ with respect to the ground plane such that  $\mathbf{g}_z=-\mathbf{r}_y$ where $\mathbf{r}_y$ is the y-axis of the camera, i.e., the rectified camera orientation is $\overline{\mathbf{R}}=\left[\begin{array}{ccc} \mathbf{g}_x^{\mathsf{T}} & -\mathbf{g}_z^{\mathsf{T}} & \mathbf{g}_y^{\mathsf{T}}\end{array}\right]^{\mathsf{T}}$.

We warp the image $\mathcal{I}$ using a homography induced by $\overline{\mathbf{R}}$, $\overline{\mathcal{I}}=\mathcal{I}(\mathbf{K}\overline{\mathbf{R}}\mathbf{R}^{\mathsf{T}}\mathbf{K}^{\mathsf{T}}\mathbf{u})$, where $\mathbf{K}$ is the intrinsic parameter and $\mathbf{u}$ is image coordinate: the camera projection matrix is $\mathbf{P}=\mathbf{K}\mathbf{R}\left[\begin{array}{cc} \mathbf{I} & -\mathbf{C}\end{array}\right] \in \mathds{R}^{3\times 4}$. This rectified image aligns with the ground plane and instantaneous velocity, which allows us to stabilize jittery first person videos due to severe head movement~\cite{su:2016,park:2016_future} as shown in Figure~\ref{Fig:rectify}. The all vanishing lines of first person images are perfectly aligned, enabling visual learning efficient where the depth and shape of an object is highly predictable by the pixel location as discussed in Section~\ref{sec:gan}.  

\subsection{ActionTunnel}
We define ActionTunnel such that it exhibits the desired property, i.e., embedding future into the present image, through three key ingredients inherited in first person images.

\begin{figure}[t]
  \centering  
      \subfigure[Height map]{\label{Fig:ego_height}\includegraphics[height=0.15\textheight]{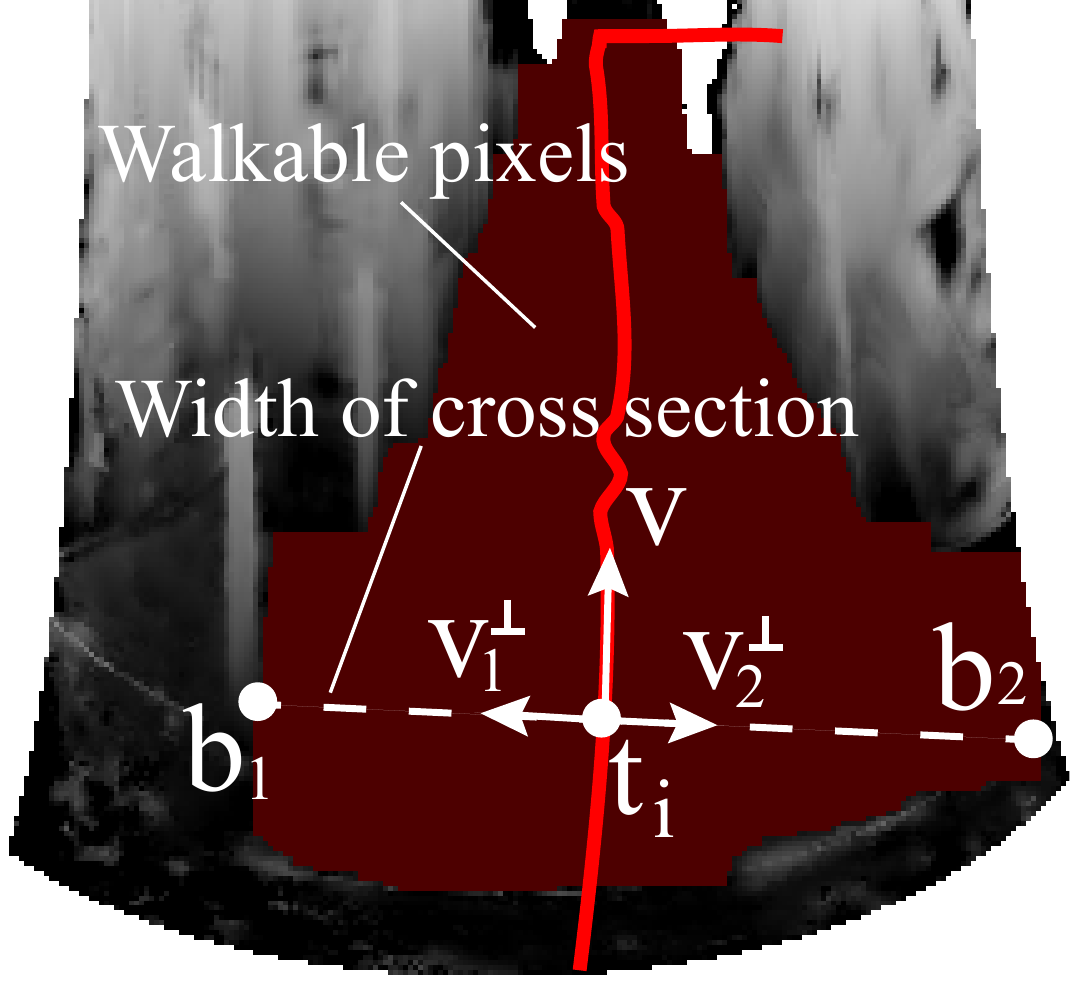}}~~~
      \subfigure[RGB map]{\label{Fig:ego_rgb}\includegraphics[height=0.15\textheight]{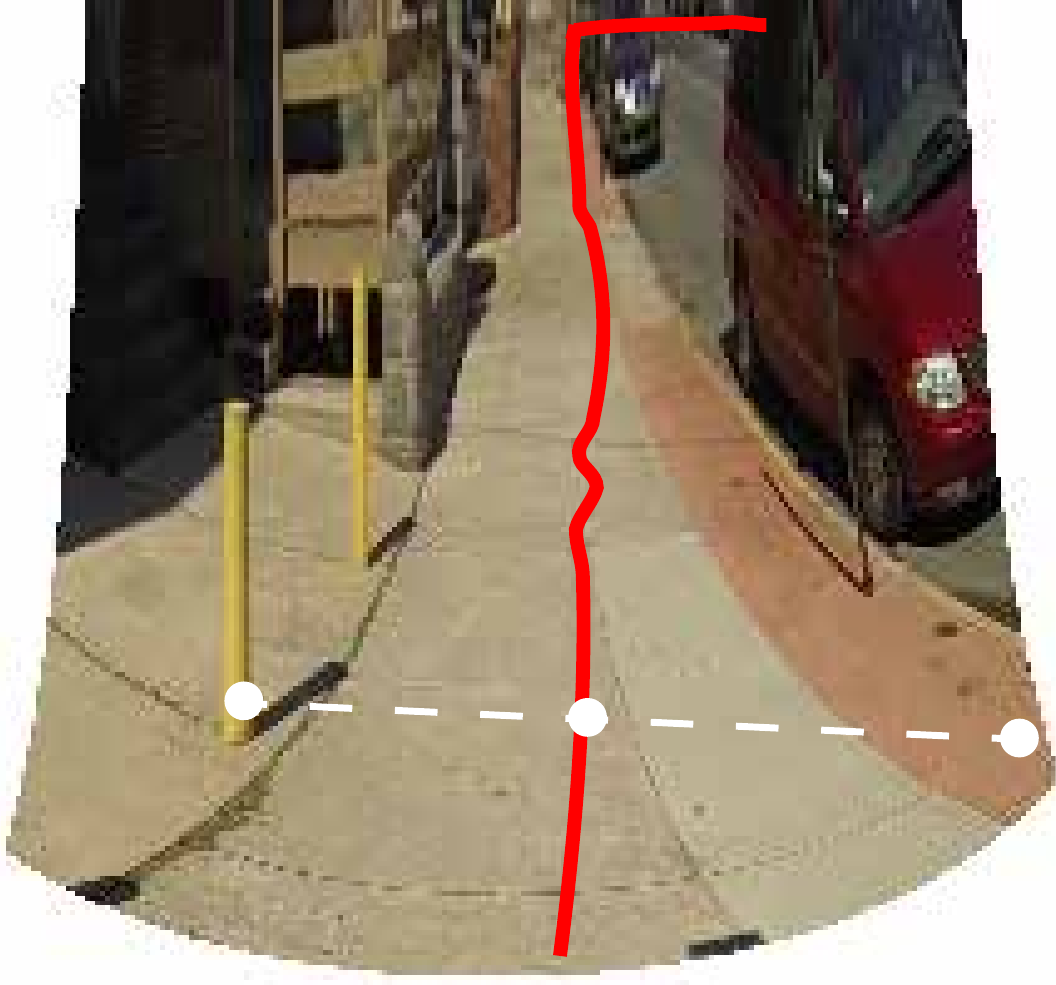}}
  \caption{We construct ActionTunnel in the 3D proxemic space (log-cylindrical coordinate). Its cross section is computed by the height map of EgoRetinal representation~\cite{park:2016_future}.} 
  \label{Fig:ego1}
\end{figure}

First, ActionTunnel encodes the action associated with a first person image. It is formed around the future trajectory that the person would walk over. We represent the future trajectory in the 3D proxemic space as $\textbf{T}=\left[\begin{array}{ccc} \mathbf{t}_1^{\mathsf{T}} & \cdots & \mathbf{t}_F^{\mathsf{T}}\end{array}\right]^{\mathsf{T}}$, where $\mathbf{t}_i=\left[\begin{array}{ccc} r_i & \theta_i & 0\end{array}\right]^{\mathsf{T}}$ is the 3D coordinates of person's foot location at the $i^{\rm th}$ time instant later in the future (Figure~\ref{Fig:ego_rgb}). 

Second, the cross section of ActionTunnel encodes the space that affords the action, i.e., sidewalk. We define the cross section as a lateral spatial extent of a walking trajectory. The lateral cross section, $\mathbf{V}=\left[\begin{array}{cccc} \mathbf{b}_1^{\mathsf{T}} & \mathbf{b}_2^{\mathsf{T}} & \mathbf{c}_1^{\mathsf{T}} & \mathbf{c}_2^{\mathsf{T}}\end{array}\right]^{\mathsf{T}}$, is defined by two bottom corners $\mathbf{b}_{1}$, $\mathbf{b}_{2}$ and two ceiling corners $\mathbf{c}_{1}$, $\mathbf{c}_{2}$ (Figure~\ref{Fig:geometry}). For the bottom corners, we seek the farthest point from the trajectory lower than $h_{\rm max}$ along $\mathbf{v}^\bot$ where $\mathbf{v}^\bot$ is the direction perpendicular to the instantaneous velocity $\mathbf{v}_i = \mathbf{t}_{i+1} - \mathbf{t}_i$ as shown in Figure~\ref{Fig:ego_height}. Formally, $\mathbf{b}_j = \lambda_{\rm max}\mathbf{v}_j^\bot + \mathbf{t}_i$ for $j=1,2$ where
\begin{align*}
    \lambda_{\rm max} = \underset{ \lambda}{\operatorname{argmax}}~~ \{\lambda | \phi(\lambda\mathbf{v}_j^\bot + \mathbf{t}_i)<h_{\rm max}, \lambda>0\}.
\end{align*}
$h_{\rm max}$ defines the walkable height in the log-cylindrical space, e.g., vehicles or buildings are not walkable. $\mathbf{v}^{\bot}$ is a vector on the ground plane perpendicular to the $\mathbf{v}_i$, i.e., $\mathbf{v}_i^{\mathsf{T}}\mathbf{v}^{\bot}=0$, and $\mathbf{v}_1^{\bot}=-\mathbf{v}_2^{\bot}$. $\phi(r,\theta,0)$ is the height map from EgoRetinal map~\cite{park:2016_future} as shown in Figure~\ref{Fig:ego_height}. The ceiling point is $\alpha H$ higher than $\mathbf{b}$ in the 3D world along $\mathbf{g}_z$ direction, which is translated to $\mathbf{c}_j=\mathbf{b}_j+\left[\begin{array}{ccc} 0 & 0 & \alpha H/Z_i\end{array}\right]^{\mathsf{T}}$ in the log-cylindrical coordinate, where $\alpha$ defines the fixed height of the ActionTunnel. 

Third, we map the texture from the first person image to the ActionTunnel surfaces defined by adjacent cross sections, $\mathcal{J}_i$. We convert the vertices of the ActionTunnel from the proxemic space to the 3D world coordinate, i.e., $X=\exp{(r)}\cos\theta$, $Y=\exp{(r)}\sin\theta$, $Z=h\exp{(r)}$, and then, project onto the rectified image plane, i.e., $\lambda\mathbf{u}=\mathbf{K}\overline{\mathbf{R}}(\mathbf{X} -\mathbf{C})$ where $\mathbf{X}=\left[\begin{array}{ccc} X & Y & Z\end{array}\right]^{\mathsf{T}}$. 

\begin{figure*}[th]
  \centering  
      \subfigure[ActionTunnel]{\label{Fig:composit1}\includegraphics[height=0.16\textheight]{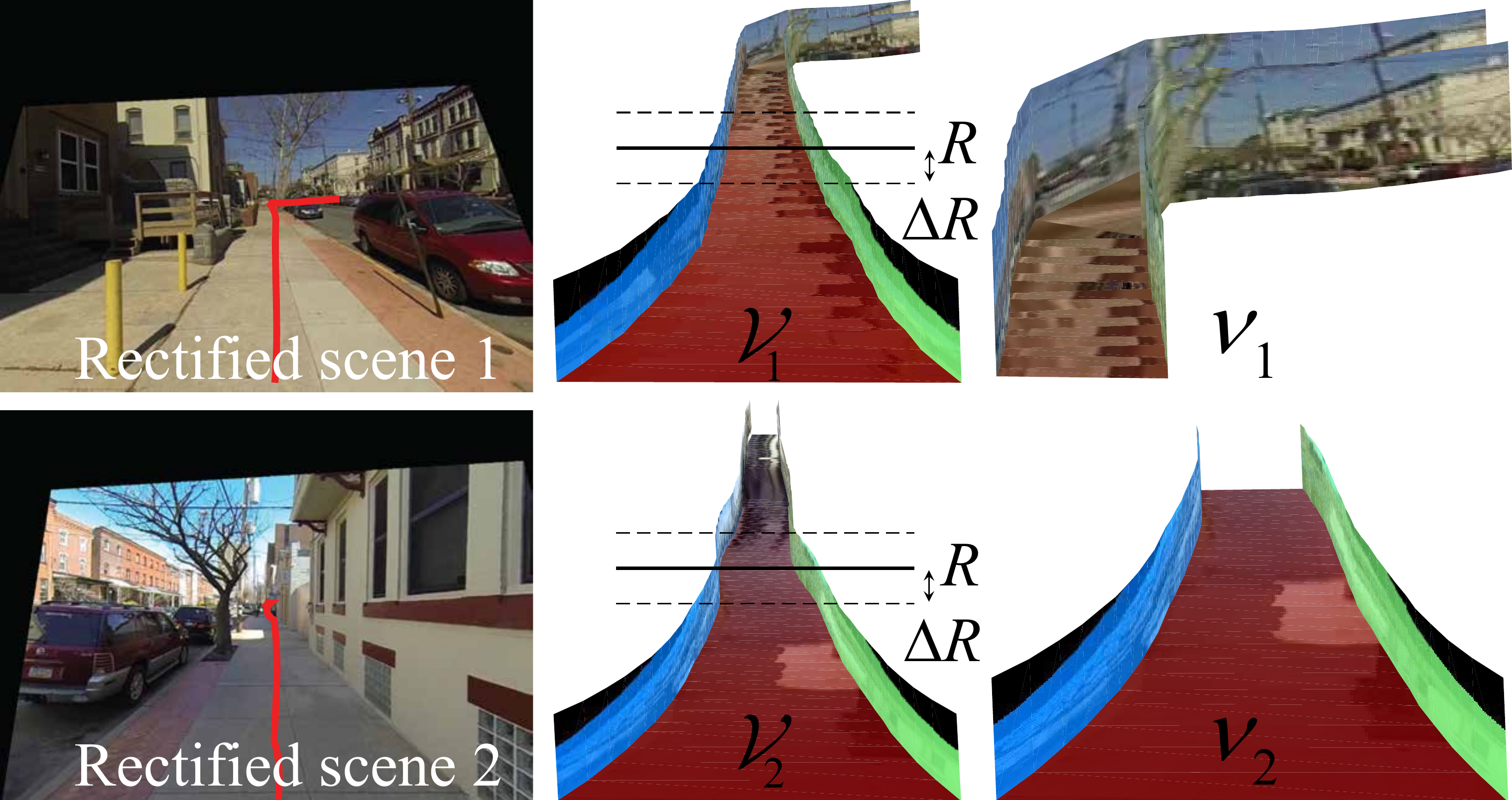}}~
      \subfigure[Connection]{\label{Fig:composit2}\includegraphics[height=0.16\textheight]{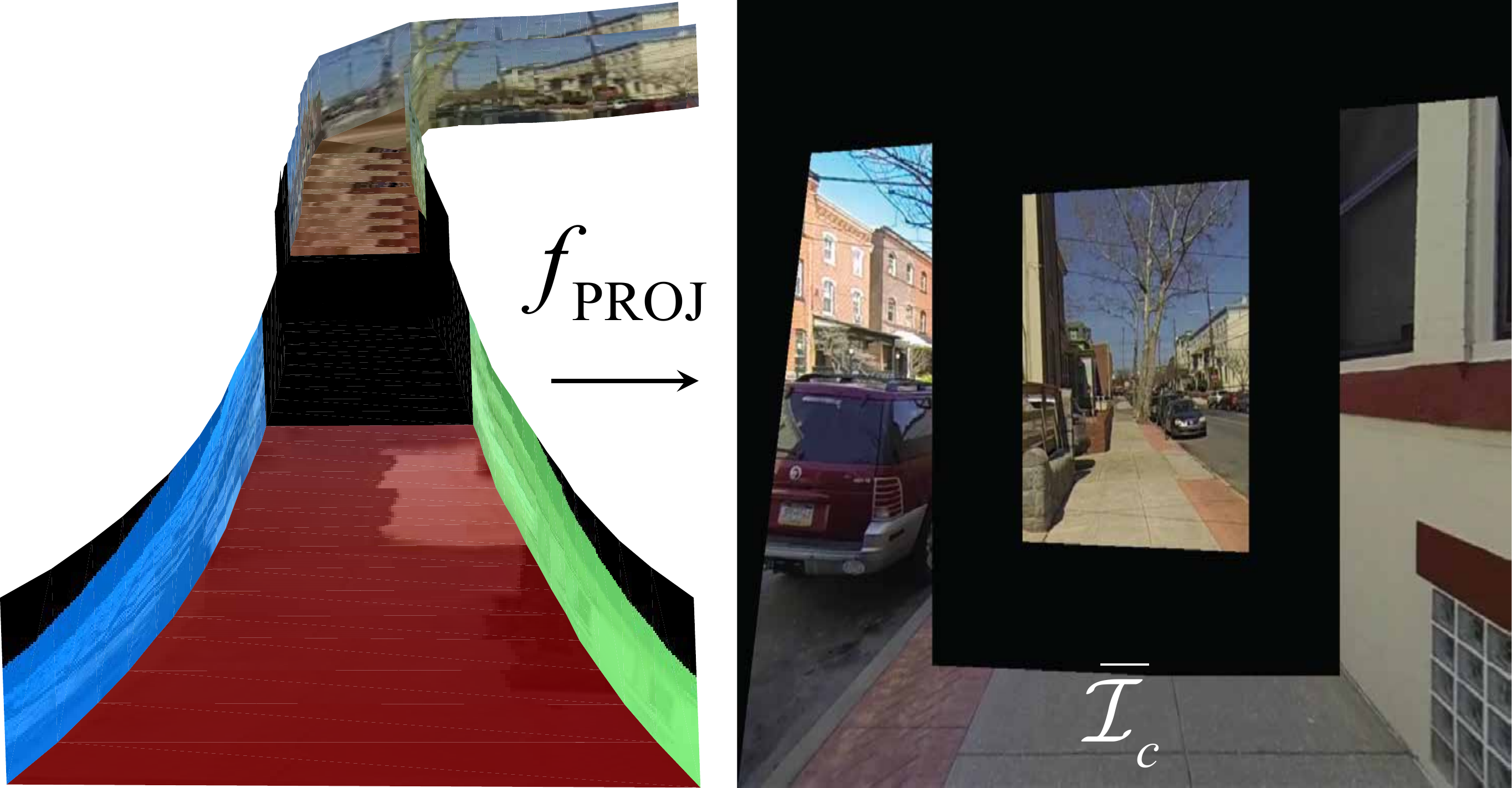}}~
      \subfigure[GAN]{\label{Fig:composit3}\includegraphics[height=0.16\textheight]{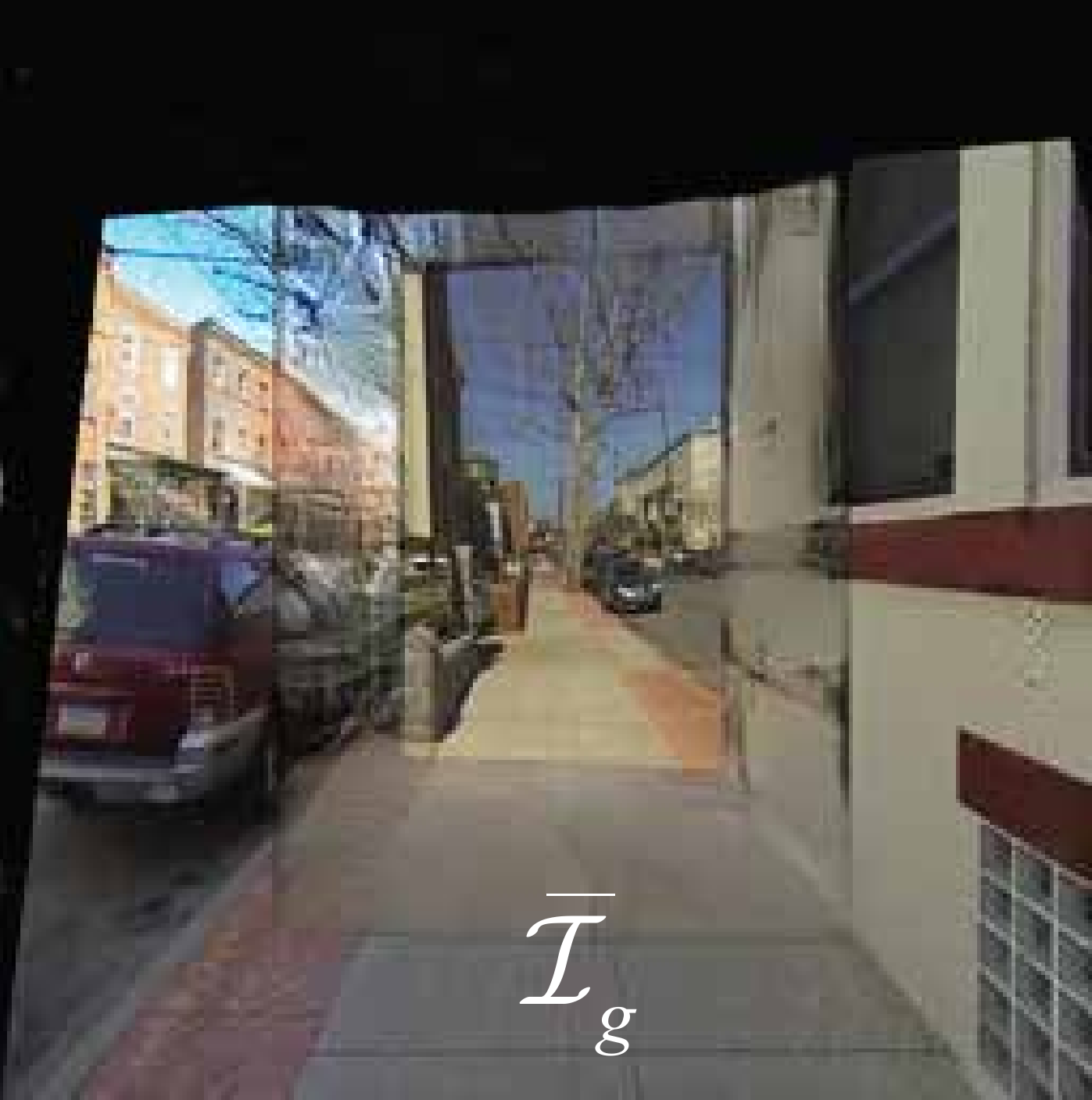}}
  \caption{(a) Given two distinctive first person scenes, (b) we construct a new ActionTunnel that defines the pixel persistence (which pixel to change). (c) We use an image generative model to complete the image.} 
  \label{Fig:composite}
\end{figure*}

In total, the ActionTunnel is composed of the series of cross section vertices and associated texture along the future trajectory $\mathcal{V}=\{\mathbf{V}_i, \mathcal{J}_i\}_{i=1}^F$. To visualize an image from a novel view, the texture of the ActionTunnel in the 3D proxemic space is projected to the rectified camera pose: $\overline{\mathcal{I}} = f_{\rm PROJ}(\mathcal{V};\overline{\mathbf{P}})$ where $\overline{\mathbf{P}} =\mathbf{K}\overline{\mathbf{R}}\left[\begin{array}{cc} \mathbf{I} & -\mathbf{C}\end{array}\right]$ is the rectified camera projection matrix, or inversely, $\mathcal{V} = f_{\rm PROJ}^{-1}(\overline{\mathcal{I}};\overline{\mathbf{P}})$ is projecting from the first person image to the ActionTunnel (Figure~\ref{Fig:composit2}). 

\section{Connecting Visual Sensations} \label{Sec:connection}
As parametrized by an action, the ActionTunnel is a medium that connects two distinctive visual sensations through walking future trajectories. In this section, we recognize a plausible transition between two visual scenes and construct a novel ActionTunnel by combining them, which allows us to create a composite picture using a generative adversarial network. We defer the visual scene retrieval to Section~\ref{Sec:ret}.

\subsection{ActionTunnel Composition} \label{Sec:composition}

Given ActionTunnel $\mathcal{V}_1=f_{\rm  PROJ}^{-1}(\overline{\mathcal{I}}_1;\overline{\mathbf{P}}_1)$ to $\mathcal{V}_2=f_{\rm  PROJ}^{-1}(\overline{\mathcal{I}}_2;\overline{\mathbf{P}}_2)$, we compose a novel transitional image, $\mathcal{I}_c$:
\begin{align}
    \overline{\mathcal{I}}_c=f_{\rm PROJ}(\mathcal{V}_c= \boldsymbol{\nu}_1 \cup \boldsymbol{\nu}_2;\overline{\mathbf{P}}_1), \label{Eq:proj}
\end{align}
where $\mathcal{V}_c$ is the new ActionTunnel constructed by combining $\boldsymbol{\nu}_1$ and $\boldsymbol{\nu}_2$. The segmented ActionTunnels are $\boldsymbol{\nu}_1 = \{\mathbf{V}_i \in \mathcal{V}_1|\mathbf{t}_i^r<R-\Delta R\}$, and $\boldsymbol{\nu}_2 = \{\mathbf{V}_i \in \mathcal{V}_2|\mathbf{t}_i^r>R+\Delta R\}$ as shown in Figure~\ref{Fig:composit1} where a transitional point $R$ with transitional length, $\Delta R$, and $\mathbf{t}^r$ is the range coordinate of trajectory point $\mathbf{t}$.
Note that $\mathcal{V}_1$ to $\mathcal{V}_2$ are aligned such that $\boldsymbol{\nu}_1 \parallel \boldsymbol{\nu}_2^\bot$ where $\boldsymbol{\nu}_2^\bot=\left(\mathcal{V}_2 \backslash \boldsymbol{\nu}_2\right)$ is the complementary set of $\boldsymbol{\nu}_2$ described in Section~\ref{Sec:ret}. Figure~\ref{Fig:composit3} illustrates the projection of the ActionTunnel onto the first person camera pose where the 3D scene is pre-aligned, e.g., vanishing lines and movement direction invariant to 3D head pitch angle. 

$\overline{\mathcal{I}}_c$ encodes the key desired property of ActionTunnel: indicating which pixel to change and stay in a spatially persistent way. The pixels in the adjoining area (masked area) are changeable while others are less changeable. 



Given $\overline{\mathcal{I}}_c$, we predict the complete image, $\overline{\mathcal{I}}_g$ including the texture of the adjoining region using a generative model $\mathsf{G}$:
\begin{align}
\overline{\mathcal{I}}_g=\mathsf{G}(\overline{\mathcal{I}}_c, \mathbf{z}; \mathbf{w}_{\mathsf{G}}) 
\end{align} 
where $\mathsf{G}$ predicts a complete image given incomplete image $\overline{\mathcal{I}}_c$ parametrized by $\mathbf{w}_{\mathsf{G}}$ as shown in Figure~\ref{Fig:composit3}. This generative model is composed of encoder $\mathsf{E}$ and decoder $\mathsf{E}^{-1}$, i.e., $\mathsf{G}=\mathsf{E}^{-1} \left(\mathsf{E}\left(\overline{\mathcal{I}}_c\right), \mathbf{z}\right)$. $\mathsf{E}$ generates a compact visual representation invariant to the location of the missing pixels~\cite{pathakCVPR16context} and $\mathsf{E}^{-1}$ reconstructs the complete image with $\mathbf{z}\sim \mathcal{N}(\mathbf{0};\mathbf{I})$ which is a random vector drawn from a zero-mean Gaussian distribution. We rotate back the complete image rectified by $\overline{\mathbf{R}}$ to the original first person camera pose by $\mathcal{I}_g=\overline{\mathcal{I}}_g(\mathbf{K}\mathbf{R}\overline{\mathbf{R}}^{\mathsf{T}}\mathbf{K}^{\mathsf{T}}\mathbf{u})$.

\subsection{Generative Image Predictor} \label{sec:gan}
We train the generative model $\mathsf{G}$ jointly with a discriminative model $\mathsf{D}$ using a generative adversarial network (GAN)~\cite{goodfellow:2014,pathakCVPR16context}. $\mathsf{D}$ differentiates a real image from an image generated by $\mathsf{G}$, while $\mathsf{G}$ is trained to confuse $\mathsf{D}$. Such adversarial setting reinforces each other, exhibiting strong predictive power. The following optimization trains these two models:
\begin{align}
 \underset{ \mathbf{w}_{\mathsf{G}}}{\operatorname{min}}~ \underset{ \mathbf{w}_{\mathsf{D}}}{\operatorname{max}}~ \mathcal{L}_{\rm GAN} + \lambda \mathcal{L}_{\rm REC},
\end{align}
where $\mathsf{D}(\overline{\mathcal{I}};\mathbf{w}_{\mathsf{D}})\in \{0,1\}$, and $\mathbf{w}_{\mathsf{D}}$ is its parameter. 

$\mathcal{L}_{\rm GAN}$ is the generative adversarial loss that measures the confusion of the discriminative network between the real and generated images: 
\begin{align}
    \mathcal{L}_{\rm GAN} = \mathbb{E}_{\overline{\mathcal{I}}}~[\log(\mathsf{D}(\overline{\mathcal{I}}))] +  \mathbb{E}_{\overline{\mathcal{I}}_c, \mathbf{z}\sim\mathcal{N}}~[\log(1-\mathsf{D}(\mathsf{G}(\overline{\mathcal{I}}_c; \mathbf{z})))]. \nonumber
\end{align}
$\mathcal{L}_{\rm REC}$ is the reconstruction error that penalizes the difference of generated image from the corresponding image:
\begin{align}
    \mathcal{L}_{\rm REC} &= \|\mathcal{M}\odot\left(\overline{\mathcal{I}}-\mathsf{G}(\overline{\mathcal{I}}_c;\mathbf{z})\right)\|_1 \nonumber
\end{align}
where $\mathcal{M}$ is a binary mask indicating the valid pixel locations in the rectified image $\overline{\mathcal{I}}$, induced by a homography $\mathbf{K}\mathbf{R}\overline{\mathbf{R}}^{\mathsf{T}}\mathbf{K}^{\mathsf{T}}$.

\begin{figure*}[th]
  \centering  
  \includegraphics[width=\textwidth]{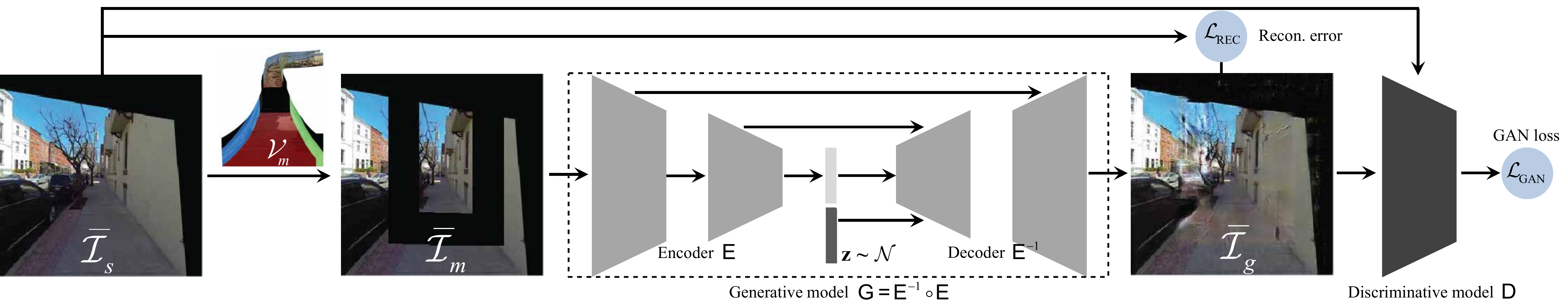}
  \caption{We leverage a generative model, $\mathsf{G}$, to predict a novel image created by ActionTunnel. We jointly train $\mathsf{G}$ with a discriminative model $\mathsf{D}$ using a generative adversarial network~\cite{goodfellow:2014}. Given a masked rectified image, $\overline{\mathcal{I}}_m$, generated by $\overline{\mathcal{I}}_s$ and $\mathcal{V}_m$, the network computes $\overline{\mathcal{I}}_g$ that minimizes reconstruction error and adversarial loss.} \label{Fig:gan}
\end{figure*}

Our GAN is inspired by the design of a deep convolutional generative adversarial network~\cite{radford:2015,pathakCVPR16context}, composed by 5 convolutional layers followed by batch normalization and Leaky ReLU as shown in Figure~\ref{Fig:gan}. The encoder takes images of size $256 \times 256 \times 3$ and outputs 100 dimensional visual feature $\mathsf{E}(\overline{\mathcal{I}}_c)$. We augment the bypassing connections in the corresponding layers of the encoder/decoder, which is beneficial for efficient training and information sharing. This allows the network to capture the details of visual structure.

\noindent\textbf{Training data} 
We generate the training data using ActionTunnel. Note that no ground truth data is available in theory. Instead, we use the same image to represent two visual scenes in Equation~(\ref{Eq:proj}). Given a rectified image $\overline{\mathcal{I}}_s$, we build a masked image, $\overline{\mathcal{I}}_m$ by constructing the ActionTunnel and leaving out vertices, i.e., $\overline{\mathcal{I}}_m = f_{\rm PROJ}(\overline{\mathcal{V}}_{m})$ where $\mathcal{V}_{m} = \{\mathbf{V}_i \in f_{\rm PROJ}^{-1}(\overline{\mathcal{I}}_s)|\mathbf{t}_i^r \notin (R-\Delta R, R+\Delta R)\}$. The training data has diverse shapes, sizes, and locations. 

As the training and testing distributions are different, i.e., testing from two images while training from the same image, it is possible that predicting $\overline{\mathcal{I}}_g$ is challenging. However, precise alignment and retrieval in Section~\ref{Sec:ret} allows us to align 3D structure resulting in plausible continuous transitions in practice.




\subsection{Retrieval} \label{Sec:ret}
Given a first person image, we find an image taken in different time and space that can produce plausible transition. Two conditions have to be made: 1) future trajectories are similar in particular, near the camera wearer, which provides an immediate continuous transition while long term future can be discounted; 2) 3D spatial layouts such as road, building, and car are aligned rather than appearance itself because motion perception is more dominant than semantic scene recognition where the camera wearer undergoes strong egomotion, e.g., we often perceive ``something'' on the left without knowing what it is when turning right: ``knowing'' vs. ``remembering''~\cite{tulving:1983}.   

We define the distance metric, $D(\cdot,\cdot)$, as follow:
\begin{align}
    D(\mathcal{I}_1, \mathcal{I}_2) = \lambda_{\rm V}D_{\rm V}(\overline{\mathcal{I}}_1, \overline{\mathcal{I}}_2) + \lambda_{\rm S}D_{\rm S}(\phi_1, \phi_2)+ \lambda_{\rm M}D_{\rm M}(\mathcal{V}_1,\mathcal{V}_2) \nonumber
\end{align}
where $D_{\rm V}$, $D_{\rm S}$, and $D_{\rm M}$ are visual, spatial, and motion distance, respectively, and $\lambda_{\rm V}$, $\lambda_{\rm S}$, and $\lambda_{\rm M}$ are their weights. 

The visual distance measures the scene difference invariant to translation: $D_{\rm V}(\overline{\mathcal{I}}_1, \overline{\mathcal{I}}_2) = \|\mathsf{F}(\overline{\mathcal{I}}_1)-\mathsf{F}(\overline{\mathcal{I}}_2)\|^2$ where $\mathsf{F}$ is a compact visual feature for the rectified image. Our choice of $\mathsf{F}$ is a convolutional neural network~\cite{krizhevsky:2012}. These visual features are highly indicative of visual scene context robust to translation. 

The spatial distance measures the 3D scene similarity which requires a precise alignment: $D_{\rm S}(\phi_1, \phi_2) =  \underset{\Delta \theta}{\operatorname{max}}~ \left(1-NCC(\phi_1(r,\Delta \theta),\phi_2(r,0))\right)$ where $NCC(\cdot,\cdot)$ measures the normalized cross-correlation between the height map ($\phi$) from EgoRetinal map. $D_{\rm S}$ finds the maximum response across angle, which produces the best alignment. 

Lastly, we measure motion difference via ActionTunnel: $D_{\rm M}= \underset{\Delta \mathbf{x}}{\operatorname{min}}~ \sum_{i=1}^F\|\mathbf{V}_i^1+\Delta \mathbf{x}-\mathbf{V}_i^2\|^2$ where $\mathbf{V}_i^1$ and $\mathbf{V}_i^2$ are the vertices of cross sections of first and second ActionTunnels, respectively, and $F$ is the number of frames in the trajectory of the ActionTunnel. This also align motion in terms of translation, $\mathbf{x}$.

Using this distance metric, we retrieve and alignment jointly. In practice, we align the angle using $D_{\rm S}$ and then, the translation using $D_{\rm M}$. This distance measure encodes two similarities: trajectory and spatial layout. 

Given $k$ nearest neighbors based on $D(\mathcal{I}_1, \mathcal{I}_2)$, we verify the plausibility of the scene transition using the discriminative model, $\mathsf{D}$, learned in Section~\ref{sec:gan}. $\mathsf{D}(\overline{\mathcal{I}}_c)$ computes the likelihood of being real images, which allows us to rank the neighbors. 

\section{Result}

\begin{figure*}[th]
  \centering  
      \includegraphics[width=\textwidth]{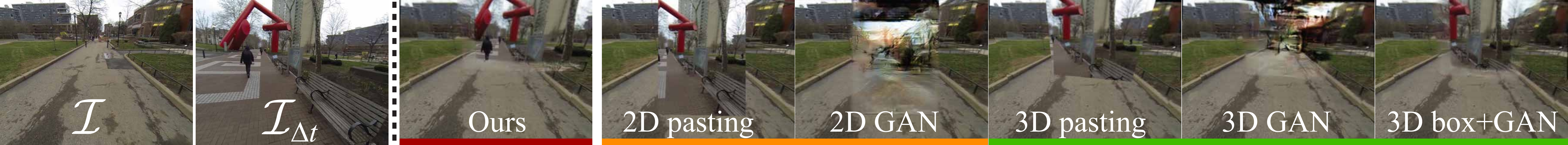}
      \subfigure[Outdoor scenes]{\label{Fig:indoor}\includegraphics[height=0.146\textheight]{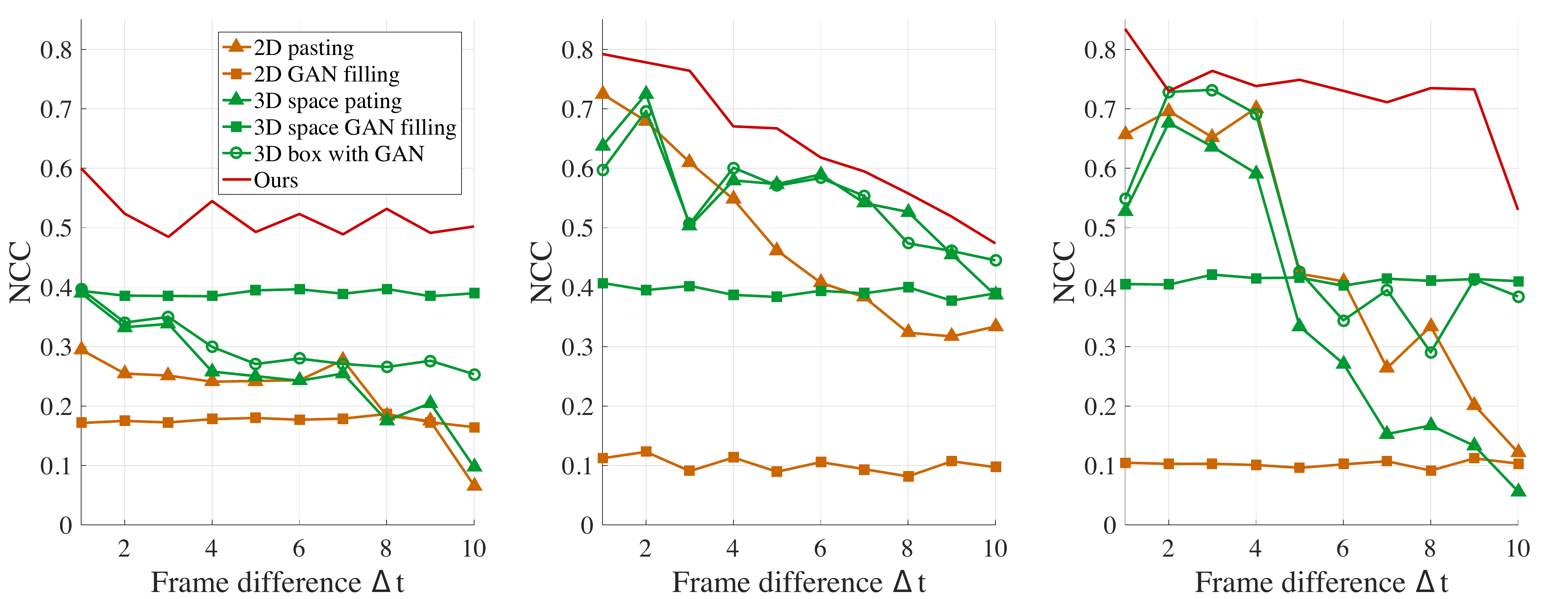}}~
      \subfigure[Indoor scenes]{\label{Fig:outdoor}\includegraphics[height=0.146\textheight]{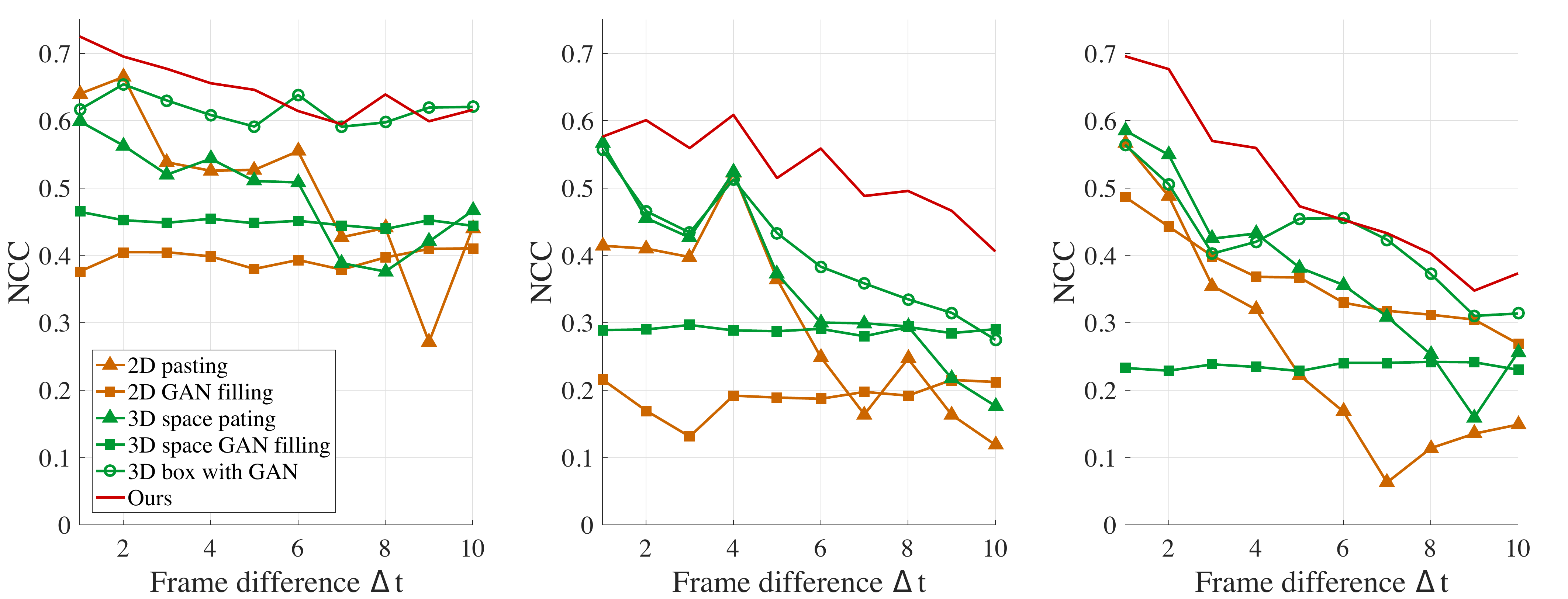}}
  \caption{We compare our method with 5 baseline algorithms for generalization power: 2D pasting, 2D GAN filling~\cite{pathakCVPR16context}, 3D space pasting, 3D space GAN filling, and 3D box with GAN~\cite{Horry:1997}. We reconstruct a masked image using a consecutive image with $\Delta t$ frame difference. Our method consistently outperforms others where stronger prediction is achieved for outdoor scenes which aligns with the prior observations~\cite{park:2016_future}.} 
  \label{Fig:quant}
\end{figure*}

We use an egomotion public dataset~\cite{park:2016_future} to evaluate our work where the 3D future trajectory per image was reconstructed using their structure from motion~\cite{hartley:2004}. This dataset includes diverse walking sequences in different locations and time (13 outdoor such as Park, Campus, and Downtown, 13 indoor scenes such as Mall, IKEA, Costco). The dataset provides the full physical scale of reconstruction at 5 fps and the disparity map from 100 mm baseline stereo system. ActionTunnel based on our 3D proxemic space in Section~\ref{Sec:fps} is computed based on their EgoRetinal representation. Note that for all evaluations, the training set is completely isolated from the testing set.




\subsection{Quantitative Evaluation} \label{Sec:quant}
As discussed in Section~\ref{sec:gan}, there exists no ground truth data because a visual scene transition is fundamentally impossible in reality. Instead, we present a unique measure that can evaluate the validity of our method: given an image, $\mathcal{I}$ with missing pixels, $\mathcal{M}_f\odot \mathcal{I}$ where the mask $\mathcal{M}_f$ indicating valid pixels, we predict them using the remaining pixels ($(1-\mathcal{M}_f) \odot \mathcal{I}$) and a consecutive image $\mathcal{I}_{\Delta t}$. We measure reconstruction validity using normalized cross correlation (NCC), i.e., $\eta = NCC(\mathcal{M}_f\odot \mathcal{I}, \mathcal{M}_f \odot \mathcal{I}_r)$ where $\mathcal{I}_r$ is the reconstructed image. This evaluation metric differs from the inpainting measure~\cite{pathakCVPR16context} because we allow additional visual cues from consecutive images. This metric can directly measure the efficacy of ActionTunnel that automatically recognizes which pixel to change and stay (pixel persistence). 

We compare our method with 5 baseline algorithms. \textbf{2D pasting}: We predict the missing pixels by copying the pixels from the consecutive frame, i.e., $(1-\mathcal{M}_f) \odot \mathcal{I} = (1-\mathcal{M}_f) \odot \mathcal{I}_{\Delta t}$. If the camera wearer does not move, this measure will be maximized. \textbf{2D GAN filling}: The prediction by the inpainting approach~\cite{pathakCVPR16context} that uses a generative adversarial network is compared. This work has been evaluated on Street view and Imagenet, which could be challenging to predict a first person image due to the larger variation of visual data caused by head movement. \textbf{3D space pasting}: We evaluate the effectiveness of knowing 3D based on our rectification using the ground plane discussed in Section~\ref{Sec:fps}. To eliminate the effect of the walking action, we do not take into account the instantaneous velocity, $\mathbf{v}$, to define $\overline{\mathbf{R}}$. We predict pixels in the rectified image by copying pixels from the consecutive image, $(1-\overline{\mathcal{M}}_f) \odot \overline{\mathcal{I}} = (1-\overline{\mathcal{M}}_f) \odot \overline{\mathcal{I}}_{\Delta t}$. Note that the mask is also rectified, $\overline{\mathcal{M}}_f$. \textbf{3D GAN space filling}: Similar to 2D GAN filling, we perform this in the rectified image where the vanishing lines of all first person images are aligned. \textbf{3D Box with GAN}: We use a 3D box representation of the world~\cite{Horry:1997} to recognize the pixel persistence. This is similar to our method except for the fact that it does not take into account actions. For each evaluation, we adjust the shape of the mask to be consistent across baseline algorithms. 

We evaluate our work in terms of two aspects: generalization power and missing data.

\noindent\textbf{Generalization power} We predict the missing pixels (more than 60\% in image) by changing $\Delta t$, the 3D distance between two images. This can evaluate how far the algorithm can benefit from the consecutive image. As increasing $\Delta t$, the visual similarity is less likely consistent. Also the head orientation plays a significant role in prediction where 3D based methods often outperform 2D based methods (2D pasting, 2D GAN filling). In particular, 2D GAN filling shows limited performance on first person image prediction due to unstable head movement. Interestingly, 2D pasting sometimes outperform 2D GAN, which indicates exploiting an additional consecutive image is highly beneficial. In contrast, 3D GAN filling produces a strong predictive power because the scene is well aligned and therefore, structured. The generalization power across $\Delta t$ reduces as it increases for 2D and 3D pasting methods while GAN based methods perform consistently as shown in Figure~\ref{Fig:quant} and Table~\ref{table:per}. Overall, our method outperforms all baselines across different scenes.

\begin{table*}[th]
\centering
\scriptsize
\begin{tabular}{l|c|c|c|c|c||c|c|c|c|c}
\hline
& \multicolumn{5}{c||}{Indoor}& \multicolumn{5}{c}{Outdoor}\\
\hline
$\Delta t$ & 2 & 4 & 6 & 8 & 10 & 2 & 4 & 6 & 8 & 10\\\hline
2D pasting&\textbf{0.55}(0.19)  &  0.40(0.18)  &  0.35(0.20) &   0.30(0.19)  &  0.23(0.19) & 0.48(0.23)  &   0.40(0.24) &    0.35(0.25)   &  0.31(0.27)  &   0.30(0.24)\\

2D GAN&0.20(0.28)   & 0.17(0.28)  &  0.18(0.25)  &  0.17(0.25) &   0.13(0.25) &    0.15(0.15)  &  0.15(0.15)&    0.15(0.15)   & 0.16(0.15) &    0.16(0.15) \\
    
3D pasting&0.53(0.16)  &  0.39(0.18)  &  0.35(0.20)   & 0.31(0.18)  &  0.26(0.19) & 0.50(0.24)  &  0.42(0.26) &   0.36(0.26)  &  0.32(0.25)    & 0.31(0.24) \\
       
3D GAN&0.29(0.14)  &  0.27(0.14)  &  0.28(0.14) &   0.28(0.15)  &  0.27(0.13)  &  0.39(0.17)  &  0.40(0.17)&    0.40(0.17) &   0.42(0.17) &   0.42(0.17)\\

3D Box+GAN&0.48(0.14)  &  0.43(0.15)  &  0.37(0.16)  &  0.35(0.14)  &  0.31(0.14) &  0.53(0.19)    & 0.49(0.20)&    0.46(0.19) &   0.45(0.19)  &  0.44(0.18) \\

Ours&0.54(0.12)  &  \textbf{0.44}(0.13)  &  \textbf{0.39}(0.13)  &  \textbf{0.37}(0.13)  &  \textbf{0.34}(0.14) & \textbf{0.60}(0.18)  &  \textbf{0.57}(0.19) &  \textbf{0.51}(0.19) &   \textbf{0.50}(0.19)&    \textbf{0.47}(0.18)\\
\hline
\end{tabular}
\caption[Normalized cross correlation (NCC) over frame difference: median(std.)]{Normalized cross correlation (NCC) over frame difference: median(std.)}
\label{table:per}
\end{table*}

\begin{figure}[h]
  \centering  
  \includegraphics[width=0.49\textwidth]{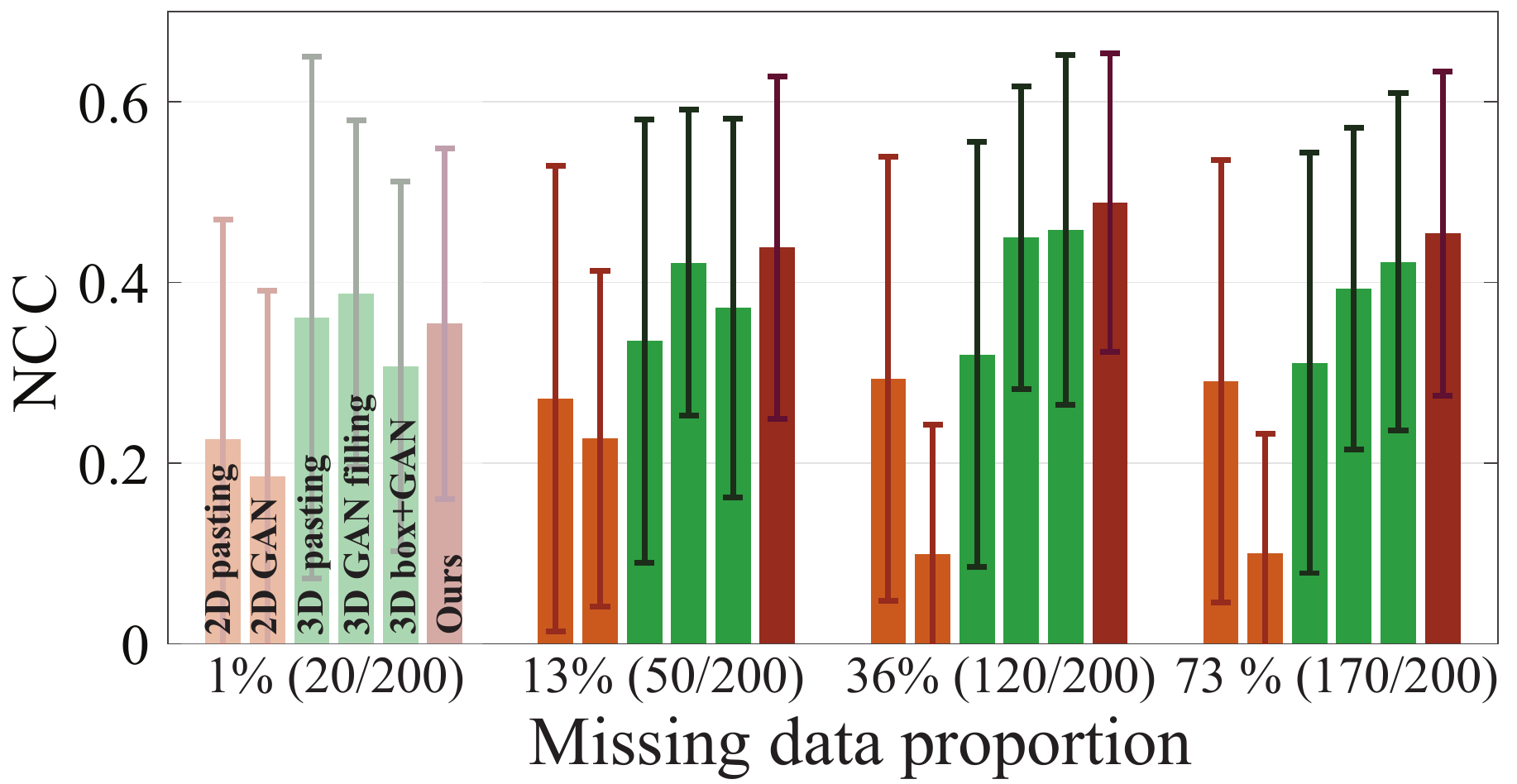}
  \caption{As missing data increases, the effect of knowing 3D information (ground plane and fugure trajectory) becomes more prominent. Our method exhibits strong predictive power over other methods.} \label{Fig:radius}
\end{figure}

\noindent\textbf{Missing data} We study the predictive power of our work as changing the proportion of missing data. The missing data is a key factor to produce a continuous spacetime travel because as the transitional point $R$ in Section~\ref{Sec:composition} approaches to the camera, the number of missing pixels increases quadratically. As shown in Figure~\ref{Fig:radius}, when the proportion of missing data is small (1\%), all methods produces similar performance. Our method shows consistent prediction regardless the missing data increases because the visuals of the consecutive frame are well aligned.


\subsection{Qualitative Evaluation}
\begin{figure*}[th]
  \centering  
      \subfigure[Outdoor: straight to right turn]{\includegraphics[width=\textwidth]{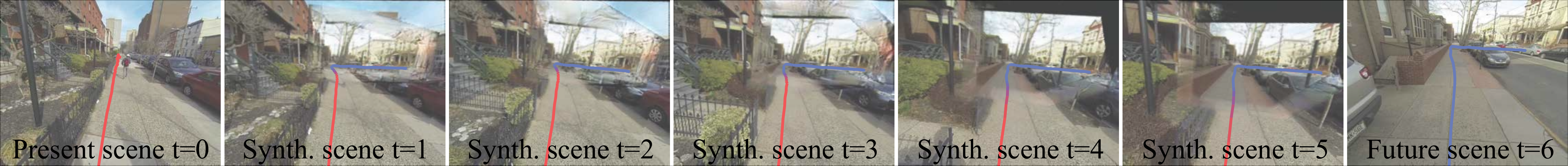}}\vspace{-2mm}
      \subfigure[Outdoor: straight to right turn]{\includegraphics[width=\textwidth]{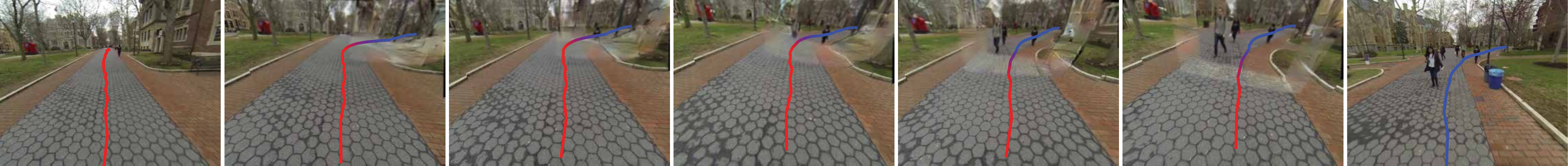}}\vspace{-2mm}
      \subfigure[Outdoor: right to left turn]{\includegraphics[width=\textwidth]{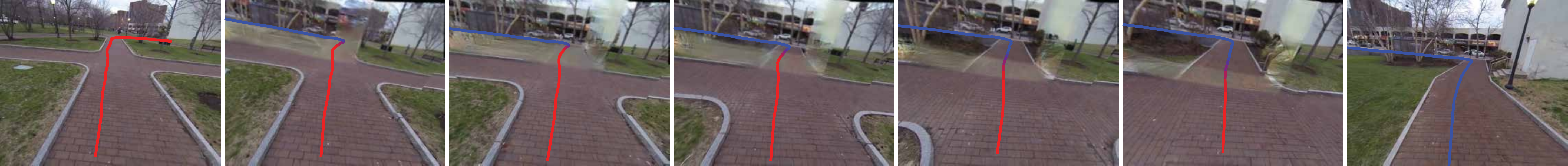}}\vspace{-2mm}
      \subfigure[Indoor (Costco): Straight to right turn]{\includegraphics[width=\textwidth]{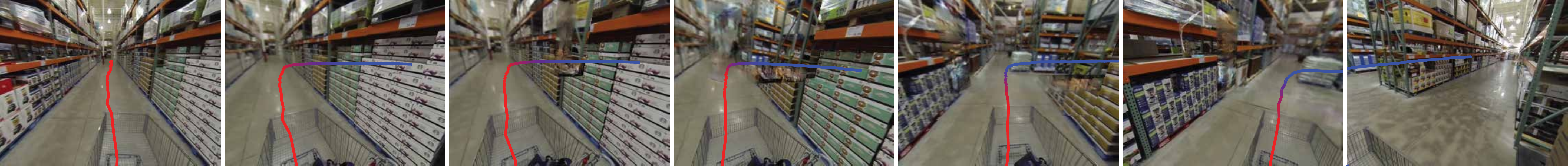}}\vspace{-2mm}
      \subfigure[Indoor (MALL): Straight to left turn]{\includegraphics[width=\textwidth]{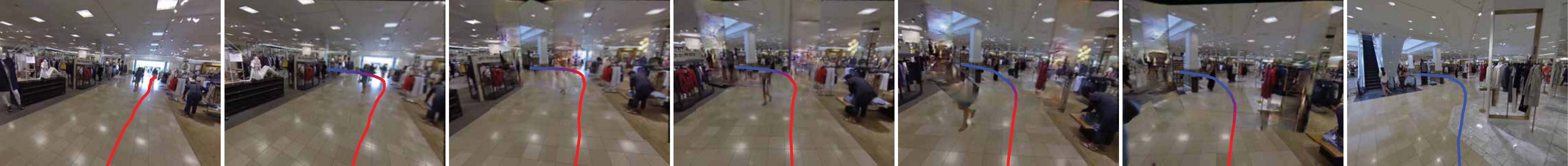}}\vspace{-2mm}
      \subfigure[Indoor (MALL): Straight to left turn]{\includegraphics[width=\textwidth]{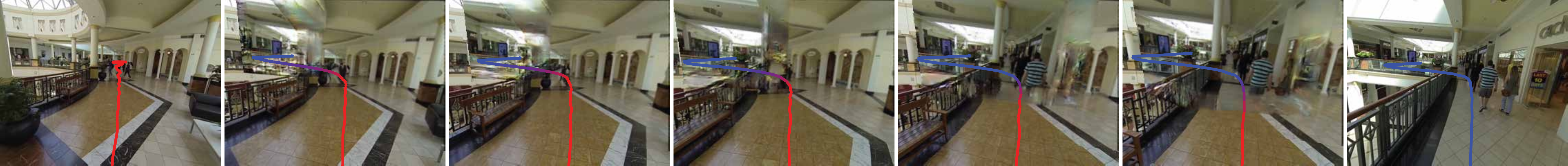}}\vspace{-2mm}
      \subfigure[Indoor (IKEA): Straight to right turn]{\includegraphics[width=\textwidth]{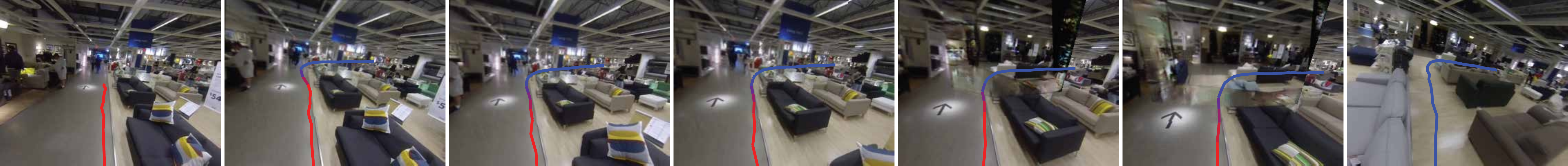}}
  \caption{We apply our method to edit the present image such that it embeds the future desired path for diverse outdoor and indoor scenes. The second column to sixth column are synthesized images over time from present scene (left) to future scene (right). The color of the trajectory encodes the transitional point between scenes. Results are best seen in color.} 
  \label{Fig:qual}
\end{figure*}

We apply our method to diverse real world scenes both outdoor and indoor. Figure~\ref{Fig:qual} illustrates a sequence of continuous transitional images from present visual scene to the future as the camera wearer walks over time. To generate the transition, we set the transitional point in 3D to compute $R$ per time instant, which allows us to travel one space to the other that affords the different future trajectory. The color of trajectory indicates the transitional point where ActionTunnels are combined. Our method produces a creative transition that generates a new space where we can cut in.

\begin{wrapfigure}{r}{0.25\textwidth}
\vspace{-8mm}
  \begin{center} 
    \includegraphics[width=0.25\textwidth]{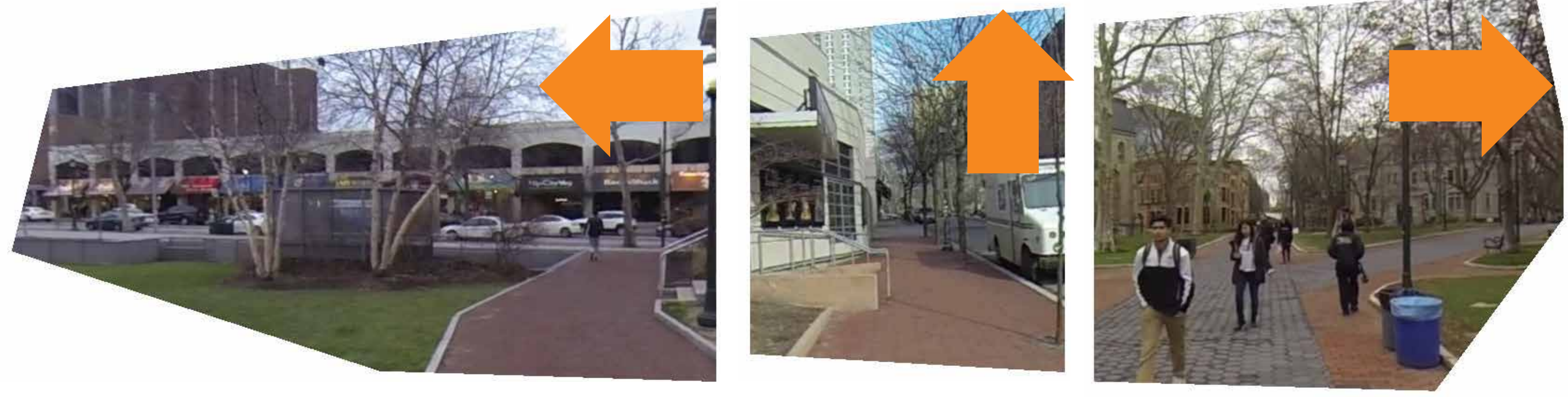}
  \end{center}
    \vspace{-7mm}

\end{wrapfigure}
The inset image on the right shows examples of visual semantics and its mask shape associated with actions (left, straight, and right) computed by ActionTunnel. Our method embeds such semantics into the present first person image to synthesize a transitional image.


\section{Summary}
This paper presents a method to synthesize a present first person image to afford the desired future action in a spatially consistent way. We leverage ActionTunnel to edit the present image to embed visual semantics of future action without manually annotation. We retrieve an image given the present scene based on similarity of future trajectories and 3D spatial layout. ActionTunnels from two images are constructed, segmented, glued, and projected onto the first person camera. The resulting image includes missing pixels where we complete using a generative adversarial network. We demonstrate that our method outperforms existing 2D based image inpainting and produces compelling visual experience by traveling through space and time.

\newpage
{\footnotesize
\bibliographystyle{ieee}
\bibliography{egbib}
}

\end{document}